\documentclass[10pt,twocolumn,letterpaper]{article}

\usepackage{cvpr}
\usepackage{times}
\usepackage{epsfig}
\usepackage{graphicx}
\usepackage{amsmath}
\usepackage{amssymb}
\usepackage{bbding}
\usepackage{subfigure}
\usepackage{color}
\usepackage[table,xcdraw]{xcolor}
\usepackage{pifont}
\usepackage{tablefootnote}

\usepackage[breaklinks=true,bookmarks=false]{hyperref}

\cvprfinalcopy 

\def\cvprPaperID{6} 

\ifcvprfinal\fi

\begin{document}

\title{The One Hundred Layers Tiramisu: \\ Fully Convolutional DenseNets for Semantic Segmentation}

\author{Simon J\'{e}gou$^1$ Michal Drozdzal$^{2,3}$ David Vazquez$^{1,4}$ Adriana Romero$^1$ Yoshua Bengio$^1$\\
$^1$Montreal Institute for Learning Algorithms $^2$\'{E}cole Polytechnique de Montr\'{e}al \\ $^3$Imagia Inc., Montr\'{e}al, $^4$Computer Vision Center, Barcelona \\
{\tt\small simon.jegou@gmail.com, michal@imagia.com, dvazquez@cvc.uab.es,} \\ {\tt \small adriana.romero.soriano@umontreal.ca, yoshua.umontreal@gmail.com}
}

\maketitle

\begin{abstract}
State-of-the-art approaches for semantic image segmentation are built on Convolutional Neural Networks (CNNs). The typical segmentation architecture is composed of (a) a downsampling path responsible for extracting coarse semantic features, followed by (b) an upsampling path trained to recover the input image resolution at the output of the model and, optionally, (c) a post-processing module (\eg Conditional Random Fields) to refine the model predictions. 

Recently, a new CNN architecture, Densely Connected Convolutional Networks (DenseNets), has shown excellent results on image classification tasks. The idea of DenseNets is based on the observation that if each layer is directly connected to every other layer in a feed-forward fashion then the network will be more accurate and easier to train.  

In this paper, we extend DenseNets to deal with the problem of semantic segmentation. We achieve state-of-the-art results on urban scene benchmark datasets such as CamVid and Gatech, without any further post-processing module nor pretraining. Moreover, due to smart construction of the model, our approach has much less parameters than currently published best entries for these datasets. Code to reproduce the experiments is publicly available here : \url{https://github.com/SimJeg/FC-DenseNet}

\end{abstract}


\section{Introduction}
\label{sec:intro} 

\begin{figure}[t!]
\centering
  \includegraphics[width=0.65\columnwidth]{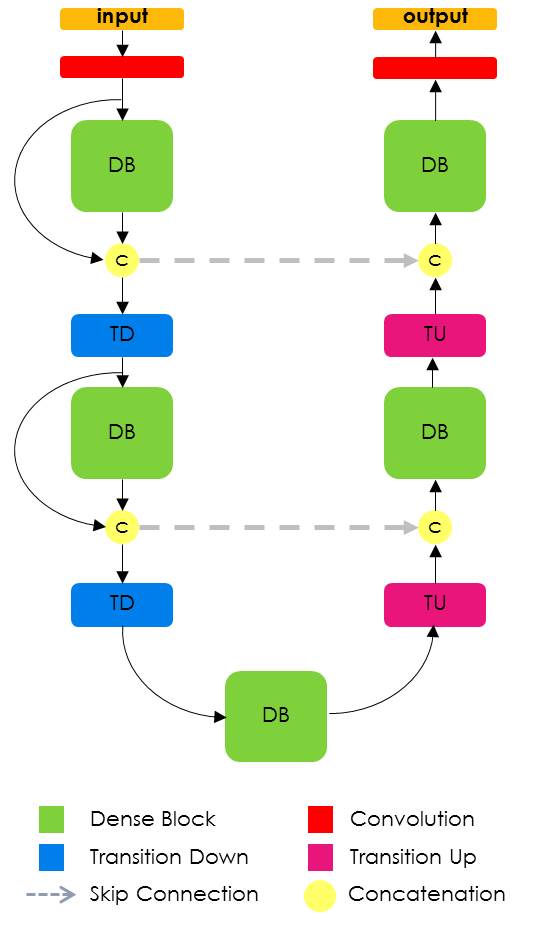}
  \label{fig:schema}	
  \caption{Diagram of our architecture for semantic segmentation. Our architecture is built from dense blocks. The diagram is composed of a downsampling path with 2 Transitions Down (TD) and an upsampling path with 2 Transitions Up (TU). A circle represents concatenation and arrows represent connectivity patterns in the network. Gray horizontal arrows represent skip connections, the feature maps from the downsampling path are concatenated with the corresponding feature maps in the upsampling path. Note that the connectivity pattern in the upsampling and the downsampling paths are different. In the downsampling path, the input to a dense block is concatenated with its output, leading to a linear growth of the number of feature maps, whereas in the upsampling path, it is not.}
\end{figure}

Convolutional Neural Networks (CNNs) are driving major advances in many computer vision tasks, such as image classification \cite{SimonyanZ14a}, object detection \cite{RenHG015,RedmonDGF15} and semantic image segmentation \cite{long2015fully}. The last few years have witnessed outstanding improvements on CNN-based models. \emph{Very deep} architectures \cite{SimonyanZ14a,HeZRS15,SzegedyLJSRAEVR14} have shown impressive results on standard benchmarks such as ImageNet \cite{imagenet_cvpr09} or MSCOCO \cite{mscoco}. State-of-the-art CNNs heavily reduce the input resolution through successive pooling layers and, thus, are well suited for applications where a single prediction per input image is expected (\eg image classification task). 

Fully Convolutional Networks (FCNs) \cite{long2015fully,ronneberger2015u} were introduced in the literature as a natural extension of CNNs to tackle per pixel prediction problems such as semantic image segmentation. FCNs add upsampling layers to standard CNNs to recover the spatial resolution of the input at the output layer. As a consequence, FCNs can process images of arbitrary size. In order to compensate for the resolution loss induced by pooling layers, FCNs introduce skip connections between their downsampling and upsampling paths. Skip connections help the upsampling path recover fine-grained information from the downsampling layers. 


Among CNN architectures extended as FCNs for semantic segmentation purposes, Residual Networks (ResNets) \cite{HeZRS15} make an interesting case. ResNets are designed to ease the training of \emph{very deep} networks (of hundreds of layers) by introducing a residual block that sums two signals: a non-linear transformation of the input and its identity mapping. The identity mapping is implemented by means of a shortcut connection. ResNets have been extended to work as FCNs \cite{ChenPK0Y16, DrozdzalVCKP16} yielding very good results in different segmentation benchmarks. ResNets incorporate additional paths to FCN (shortcut paths) and, thus, increase the number of connections within a segmentation network. This additional shortcut paths have been shown not only to improve the segmentation accuracy but also to help the network optimization process, resulting in faster convergence of the training \cite{DrozdzalVCKP16}.

Recently, a new CNN architecture, called \emph{DenseNet}, was introduced in \cite{DenseNet2016}. DenseNets are built from \emph{dense blocks} and pooling operations, where each dense block is an iterative concatenation of previous feature maps. This architecture can be seen as an extension of ResNets \cite{HeZRS15}, which performs iterative summation of previous feature maps. However, this small modification has some interesting implications: (1) parameter efficiency, DenseNets are more efficient in the parameter usage; (2) implicit deep supervision, DenseNets perform deep supervision thanks to short paths to all feature maps in the architecture (similar to Deeply Supervised Networks \cite{LeeXGZT15}); and (3) feature reuse, all layers can easily access their preceding layers making it easy to reuse the information from previously computed feature maps. The characteristics of DenseNets make them a \emph{very good fit} for semantic segmentation as they naturally induce skip connections and multi-scale supervision. 

In this paper, we extend DenseNets to work as FCNs by adding an upsampling path to recover the full input resolution. Naively building an upsampling path would result in a \emph{computationally intractable} number of feature maps with very high resolution prior to the softmax layer. This is because one would multiply the high resolution feature maps with a large number of input filters (from all the layers below), resulting in both very large amount of computation and number of parameters. In order to mitigate this effect, we \emph{only} upsample the feature maps created by the preceding dense block. Doing so allows to have a number of dense blocks at each resolution of the upsampling path independent of the number of pooling layers. Moreover, given the network architecture, the upsampled dense block combines the information contained in the other dense blocks of the same resolution. The higher resolution information is passed by means of a standard skip connection between the downsampling and the upsampling paths. The details of the proposed architecture are shown in Figure \ref{fig:schema}. We evaluate our model on two challenging benchmarks for urban scene understanding, Camvid \cite{camvid} and Gatech \cite{Gatech}, and confirm the potential of DenseNets for semantic segmentation by improving the state-of-the-art. 

Thus, the contributions of the paper can be summarized as follows: 
\begin{itemize}
\item We carefully extend the DenseNet architecture \cite{DenseNet2016} to fully convolutional networks for semantic segmentation, while mitigating the feature map explosion.
\item We highlight that the proposed upsampling path, built from dense blocks, performs better than upsampling path with more standard operations, such as the ones in \cite{ronneberger2015u}.
\item We show that such a network can outperform current state-of-the-art results on standard benchmarks for urban scene understanding without neither using pre-trained parameters nor any further post-processing.
\end{itemize}

\section{Related Work}
\label{sec:related_work} 

Recent advances in semantic segmentation have been devoted to improve architectural designs by (1) improving the upsampling path and increasing the connectivity within FCNs \cite{ronneberger2015u, SegNet2015, noh2015learning, DrozdzalVCKP16}; (2) introducing modules to account for broader context understanding \cite{VisinKCBMC15, chen14semantic, YuKoltun2016}; and/or (3) endowing FCN architectures with the ability to provide structured outputs \cite{Koltun11, chen14semantic, CRFasRNN}. 

First, different alternatives have been proposed in the literature to address the resolution recovery in FCN's upsampling path; from simple bilinear interpolation \cite{Gatta14-deepvision, long2015fully, SegNet2015} to more sophisticated operators such as unpooling \cite{SegNet2015, noh2015learning} or transposed convolutions \cite{long2015fully}. Skip connections from the downsampling to the upsampling path have also been adopted to allow for a finer information recovery \cite{ronneberger2015u}. More recently, \cite{DrozdzalVCKP16} presented a thorough analysis on the combination of identity mapping \cite{HeZRS15} and long skip connections \cite{ronneberger2015u} for semantic segmentation.

Second, approaches that introduce larger context to semantic segmentation networks include \cite{Gatta14-deepvision, VisinKCBMC15, chen14semantic, YuKoltun2016}. In \cite{Gatta14-deepvision}, an unsupervised global image descriptor is computed added to the feature maps for each pixel. In \cite{VisinKCBMC15}, Recurrent Neural Networks (RNNs) are used to retrieve contextual information by sweeping the image horizontally and vertically in both directions. In \cite{chen14semantic}, dilated convolutions are introduced as an alternative to late CNN pooling layers to capture larger context without reducing the image resolution. Following the same spirit, \cite{YuKoltun2016} propose to provide FCNs with a context module built as a stack of dilated convolutional layers to enlarge the field of view of the network. 

\begin{figure}[t!]
\centering
  \includegraphics[width=0.2\textwidth]{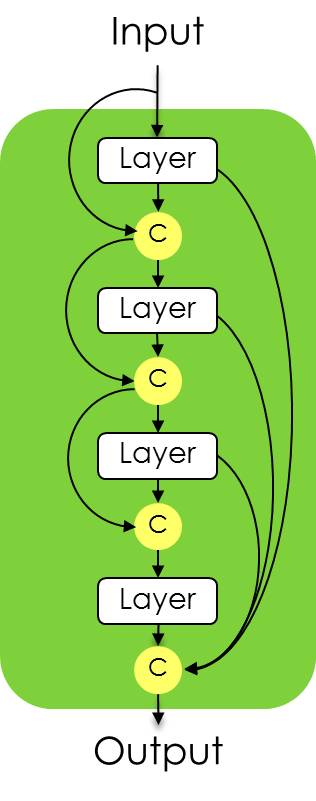}
  \caption{\label{fig:DenseBlock}Diagram of a dense block of 4 layers. A first layer is applied to the input to create $k$ feature maps, which are concatenated to the input. A second layer is then applied to create another $k$ features maps, which are again concatenated to the previous feature maps. The operation is repeated 4 times. The output of the block is the concatenation of the outputs of the 4 layers, and thus contains $4*k$ feature maps}
\end{figure}

Third, Conditional Random Fields (CRF) have long been a popular choice to enforce structure consistency to segmentation outputs. More recently, fully connected CRFs \cite{Koltun11} have been used to include structural properties of the output of FCNs \cite{chen14semantic}. Interestingly, in \cite{CRFasRNN}, RNN have been introduced to approximate mean-field iterations of CRF optimization, allowing for an end-to-end training of both the FCN and the RNN.

Finally, it is worth noting that current state-of-the-art FCN architectures for semantic segmentation often rely on pre-trained models (\eg VGG \cite{SimonyanZ14a} or ResNet101 \cite{HeZRS15}) to improve their segmentation results \cite{long2015fully, SegNet2015, ChenPK0Y16}.

\section{Fully Convolutional DenseNets}
\label{sec:dense_nets}
As mentioned in Section \ref{sec:intro}, FCNs are built from a downsampling path, an upsampling path and skip connections. Skip connections help the upsampling path recover spatially detailed information from the downsampling path, by reusing features maps. The goal of our model is to further exploit the feature reuse by extending the more sophisticated DenseNet architecture, while avoiding the feature explosion at the upsampling path of the network.

In this section, we detail the proposed model for semantic segmentation. First, we review the recently proposed DenseNet architecture. Second, we introduce the construction of the novel upsampling path and discuss its advantages w.r.t. a naive DenseNet extension and more classical architectures. Finally, we wrap up with the details of the main architecture used in Section \ref{sec:experiments}.

\subsection{Review of DenseNets}
\label{ssec:densenetreview}

Let $x_\ell$ be the output of the $\ell^{th}$ layer. In a standard CNN, $x_\ell$ is computed by applying a non-linear transformation $H_\ell$ to the output of the previous layer $x_{\ell-1}$
\begin{equation}
x_\ell = H_\ell(x_{\ell-1}),
\end{equation}
where $H$ is commonly defined as a convolution followed by a rectifier non-linearity (ReLU) and often dropout  \cite{srivastava14a}.

In order to ease the training of very deep networks, ResNets \cite{HeZRS15} introduce a residual block that sums the identity mapping of the input to the output of a layer. The resulting output $x_{\ell}$ becomes
\begin{equation}
x_\ell = H_\ell(x_{\ell-1}) + x_{\ell-1}, 
\end{equation}
allowing for the reuse of features and permitting the gradient to flow directly to earlier layers. In this case, $H$ is defined as the repetition (2 or 3 times) of a block composed of Batch Normalization (BN) \cite{IoffeS15}, followed by ReLU and a convolution. 

Pushing this idea further, DenseNets \cite{DenseNet2016} design a more sophisticated connectivity pattern that iteratively concatenates all feature outputs in a feedforward fashion. Thus, the output of the $\ell^{th}$ layer is defined as
\begin{equation}
x_\ell = H_\ell([x_{\ell-1},x_{\ell-2},...,x_0]),
\end{equation}
where $[\,... \,]$ represents the concatenation operation. In this case, $H$ is defined as BN, followed by ReLU, a convolution and dropout. Such connectivity pattern strongly encourages the reuse of features and makes all layers in the architecture receive direct supervision signal. The output dimension of each layer $\ell$ has $k$ feature maps where $k$, hereafter referred as to \emph{growth rate} parameter, is typically set to a small value (\eg $k=12$). Thus, the number of feature maps in DenseNets grows linearly with the depth (\eg after $\ell$ layers, the input $[x_{\ell-1},x_{\ell-2},...,x_0]$ will have $\ell \times k$ feature maps). 

A \emph{transition down} is introduced to reduce the spatial dimensionality of the feature maps. Such transformation is composed of a $1\times1$ convolution (which conserves the number of feature maps) followed by a $2\times2$ pooling operation. 

In the remainder of the article, we will call \emph{dense block} the concatenation of the \emph{new} feature maps created at a given resolution. Figure \ref{fig:DenseBlock} shows an example of dense block construction. Starting from an input $x_0$ (input image or output of a transition down) with $m$ feature maps, the first layer of the block generates an output $x_1$ of dimension $k$ by applying $H_1(x_0)$. These $k$ feature maps are then stacked to the previous $m$ feature maps by concatenation ($[x_1, x_0]$) and used as input to the second layer. The same operation is repeated $n$ times, leading to a new dense block with $n \times k$ feature maps.

\begin{table*}[!htb]
    \begin{minipage}{.3\linewidth}
      \centering
        \begin{tabular}{| c |}
\hline 
\textbf{Layer} \\ \hline \hline
Batch Normalization \\ \hline 
ReLU \\ \hline 
$3\times3$ Convolution \\ \hline Dropout $p=0.2$ \\ \hline   
\multicolumn{1}{c}{} \\
\end{tabular}
    \end{minipage}%
    \begin{minipage}{.3\linewidth}
      \centering
\begin{tabular}{| c |}
\hline  
\textbf{Transition Down (TD)} \\ \hline \hline 
Batch Normalization \\ \hline 
ReLU \\ \hline 
$1\times1$ Convolution  \\ \hline 
Dropout  $p=0.2$ \\ \hline  
$2\times2$ Max Pooling \\ \hline  
\multicolumn{1}{c}{} \\
\end{tabular}
    \end{minipage} 
     \begin{minipage}{.3\linewidth}
      \centering
        \begin{tabular}{| c |}
\hline  
\textbf{Transition Up (TU)} \\ \hline \hline 
$3\times3$ Transposed Convolution \\ $stride=2$   \\ \hline
\multicolumn{1}{c}{} \\
\end{tabular}
    \end{minipage}%
\caption{Building blocks of fully convolutional DenseNets. From left to right: layer used in the model, Transition Down (TD) and Transition Up (TU). See text for details.}
\label{tab:blocks}
\end{table*}

\subsection{From DenseNets to Fully Convolutional DenseNets}
\label{ssec:fcdensenets}

The DenseNet architecture described in Subsection \ref{ssec:densenetreview} constitutes the downsampling path of our Fully Convolutional DenseNet (FC-DenseNet). Note that, in the downsampling path, the linear growth in the number of features is compensated by the reduction in spatial resolution of each feature map after the pooling operation. The last layer of the downsampling path is referred to as \emph{bottleneck}.

In order to recover the input spatial resolution, FCNs introduce an upsampling path composed of convolution, upsampling operations (transposed convolutions or unpooling operations) and skip connections. In FC-DenseNets, we substitute the convolution operation by a dense block and an upsampling operation referred to as \emph{transition up}. Transition up modules consist of a transposed convolution that upsamples the previous feature maps. The upsampled feature maps are then concatenated to the ones coming from the skip connection to form the input of a new dense block. Since the upsampling path increases the feature maps spatial resolution, the linear growth in the number of features would be too memory demanding, especially for the full resolution features in the pre-softmax layer.

In order to overcome this limitation, the input of a dense block is not concatenated with its output. Thus, the transposed convolution is applied only to the feature maps obtained by \emph{the last dense block and not to all feature maps concatenated so far}. The last dense block summarizes the information contained in all the previous dense blocks at the same resolution. Note that some information from earlier dense blocks is lost in the transition down due to the pooling operation. Nevertheless, this information is available in the downsampling path of the network and can be passed via skip connections. Hence, the dense blocks of the upsampling path are computed using all the available feature maps at a given resolution. Figure \ref{fig:schema} illustrates this idea in detail.

Therefore, our upsampling path approach allows us to build very deep FC-DenseNets without a feature map explosion. An alternative way of implementing the upsampling path would be to perform consecutive transposed convolutions and complement them with skip connections from the downsampling path in a U-Net \cite{ronneberger2015u} or FCN-like \cite{long2015fully} fashion. This will be further discussed in Section \ref{sec:experiments}

\subsection{Semantic Segmentation Architecture}
\label{ssec:architecture}

\begin{table}[!b]
    \begin{minipage}{\linewidth}
      \centering
      \begin{tabular}{| c |}
      \hline 
  	  \multicolumn{1}{|c|}{\textbf{Architecture}}\\ \hline   \hline 
      \multicolumn{1}{|c|}{Input, $m=3$} \\ \hline 
      \multicolumn{1}{|c|}{$3\times3$ Convolution, $m=48$}\\ \hline 
  
      DB (4 layers) + TD, $m=112$ \\ \hline 
      DB (5 layers) + TD, $m=192$ \\ \hline 
      DB (7 layers) + TD, $m=304$ \\ \hline 
      DB (10 layers) + TD, $m=464$ \\ \hline 
      DB (12 layers) + TD, $m=656$ \\ \hline 
      \multicolumn{1}{|c|}{DB (15 layers), $m=896$} \\ \hline 
  
      TU + DB (12 layers), $m=1088$ \\ \hline 
      TU + DB (10 layers), $m=816$ \\\hline 
      TU + DB (7 layers), $m=578$ \\ \hline 
      TU + DB (5 layers), $m=384$ \\ \hline 
      TU + DB (4 layers), $m=256$\\ \hline
      \multicolumn{1}{|c|}{$1\times1$ Convolution, $m=c$} \\ \hline 
      \multicolumn{1}{|c|}{Softmax} \\ \hline 
      \end{tabular}
    \end{minipage}%
    \vspace{0.2cm}
\caption{Architecture details of FC-DenseNet103 model used in our experiments. This model is built from 103 convolutional layers. In the Table we use following notations: DB stands for Dense Block, TD stands for Transition Down, TU stands for Transition Up, BN stands for Batch Normalization and $m$ corresponds to the total number of feature maps at the end of a block. $c$ stands for the number of classes.}
\label{tab:architecture}
\end{table}

In this subsection, we detail the main architecture, \textit{FC-DenseNet103}, used in Section \ref{sec:experiments}. 

First, in Table \ref{tab:blocks}, we define the dense block layer, transition down and transition up of the architecture. Dense block layers are composed of BN, followed by ReLU, a $3 \times 3$ same convolution (no resolution loss) and dropout with probability $p=0.2$. The growth rate of the layer is set to $k=16$. Transition down is composed of BN, followed by ReLU, a $1 \times 1$ convolution, dropout with $p=0.2$ and a non-overlapping max pooling of size $2 \times 2$. Transition up is composed of a $3 \times 3$ transposed convolution with stride 2 to compensate for the pooling operation.

Second, in Table \ref{tab:architecture}, we summarize all Dense103 layers. This architecture is built from 103 convolutional layers : a first one on the input, 38 in the downsampling path, 15 in the bottleneck and 38 in the upsampling path. We use 5 Transition Down (TD), each one containing one extra convolution, and 5 Transition Up (TU), each one containing a transposed convolution. The final layer in the network is a $1\times1$ convolution followed by a softmax non-linearity to provide the per class distribution at each pixel. 

It is worth noting that, as discussed in Subsection \ref{ssec:fcdensenets}, the proposed upsampling path properly mitigates the DenseNet feature map explosion, leading to reasonable pre-softmax feature map number of 256.

Finally, the model is trained by minimizing the pixel-wise cross-entropy loss.

\section{Experiments}
\label{sec:experiments} 
We evaluate our method on two urban scene understanding datasets: CamVid \cite{camvid}, and Gatech \cite{Gatech}. We trained our models \emph{from scratch without using any extra-data nor post-processing module}. We report the results using the Intersection over Union (IoU) metric and the global accuracy (pixel-wise accuracy on the dataset). For a given class $c$, predictions $(o_i)$ and targets $(y_i)$, the IoU is defined by 
\begin{equation}
IoU(c) = \frac{\sum_i{(o_i==c \land y_i==c)}}{\sum_i{(o_i==c \lor y_i==c)}},
\end{equation}
where $\land$ is a logical \textit{and} operation, while $\lor$ is a logical \textit{or} operation. We compute $IoU$ by summing over all the pixels $i$ of the dataset.

\subsection{Architecture and training details}
\label{ssec:details}

We initialize our models using HeUniform \cite{HeZR015} and train them with RMSprop \cite{rmsprop}, with an initial learning rate of $1e-3$ and an exponential decay of 0.995 after each epoch. All models are trained on data augmented with random crops and vertical flips. For all experiments, we finetune our models with full size images and learning rate of $1e-4$. We use validation set to earlystop the training and the finetuning. We monitor mean IoU or mean accuracy and use patience of 100 (50 during finetuning). 

We regularized our models with a weight decay of $1e-4$ and a dropout rate of $0.2$. For batch normalization, we use current batch statistics at training, validation and test time.

\subsection{CamVid dataset}

CamVid\footnote{http://mi.eng.cam.ac.uk/research/projects/VideoRec/CamVid/} \cite{camvid} is a dataset of fully segmented videos for urban scene understanding. We used the split and image resolution from \cite{SegNet2015}, which consists of 367 frames for training, 101 frames for validation and 233 frames for test. Each frame has a size $360 \times 480$ and its pixels are labeled with 11 semantic classes. Our models were trained with crops of $224 \times 224$ and batch size 3. At the end, the model is finetuned with full size images.

In Table \ref{tab:CamVid}, we report our results for three networks with respectively (1) 56 layers (\emph{FC-DenseNet56}), with 4 layers per dense block and a growth rate of $12$; (2) 67 layers (\emph{FC-DenseNet67}) with 5 layers per dense block and a growth rate of $16$; and (3) 103 layers (\emph{FC-DenseNet103}) with a growth rate $k=16$ (see Table \ref{tab:architecture} for details). We also trained an architecture using standard convolutions in the upsampling path instead of dense blocks (\emph{Classic Upsampling}). In the latter architecture, we used 3 convolutions per resolution level with respectively 512, 256, 128, 128 and 64 filters, as in \cite{ronneberger2015u}. Results show clear superiority of the proposed upsampling path w.r.t. the classic one, consistently improving the IoU significantly for all classes. Particularly, we observe that unrepresented classes benefit notably from the FC-DenseNet architecture, namely \emph{sign}, \emph{pedestrian}, \emph{fence}, \emph{cyclist} experience a crucial boost in performance (ranging from $15\%$ to $25\%$).

As expected, when comparing FC-DenseNet56 or FC-DenseNet67 to FC-DenseNet103, we see that the model benefits from having more depth as well as more parameters. 

When compared to other methods, we show that FC-DenseNet architectures achieve state-of-the-art, improving upon models with 10 times more parameters. It is worth mentioning that our small model FC-DenseNet56 already outperforms popular architectures with at least 100 times more parameters. 

It is worth noting that images in CamVid correspond to video frames and, thus, the dataset contains temporal information. Some state-of-the-art methods such as \cite{KunduCVPR16} incorporate long range spatio-temporal regularization to the output of a FCN to boost their performance. Our model is able to outperform such state-of-the-art model, without requiring any temporal smoothing. However, any post-processing temporal regularization is complementary to our approach and could bring additional improvements.

Unlike most of the current state-of-the-art methods, FC-DenseNets have not been pretrained on large datasets such as ImageNet \cite{imagenet_cvpr09} and could most likely benefit from such pretraining. More recently, it has been shown that deep networks can also boost their performance when pretrained on data other than natural images, such as video games \cite{Richter_2016_ECCV,Ros_2016_CVPR} or clipart \cite{CastrejonAVPT16}, and this an interesting direction to explore.

\begin{figure*}[t!]
\centering
\subfigure{\includegraphics[width=0.9\textwidth]{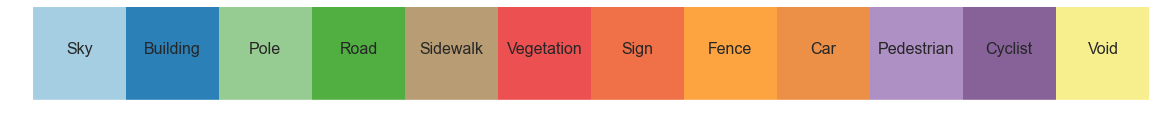}\label{fig:legend}}\hfill 
\subfigure{\includegraphics[width=0.27\textwidth]{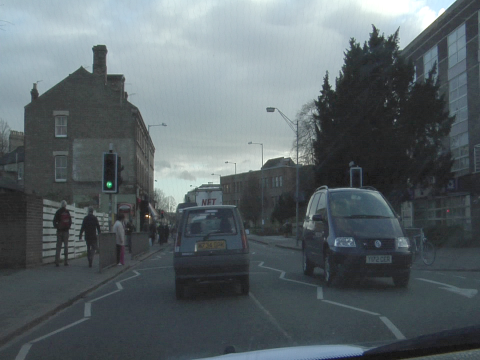}\label{fig:img1}}\hfill
\subfigure{\includegraphics[width=0.27\textwidth]{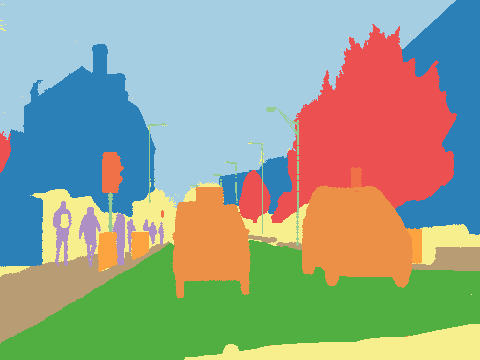}\label{fig:GT1}}\hfill 
\subfigure{\includegraphics[width=0.27\textwidth]{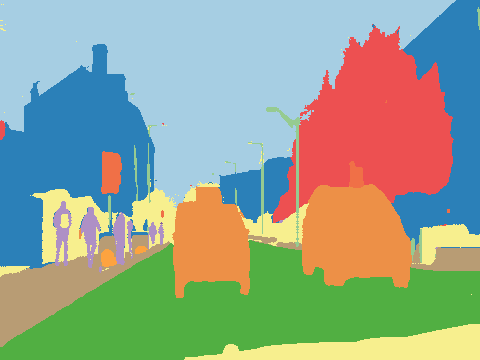}\label{fig:pred1}}\hfill \\
\subfigure{\includegraphics[width=0.27\textwidth]{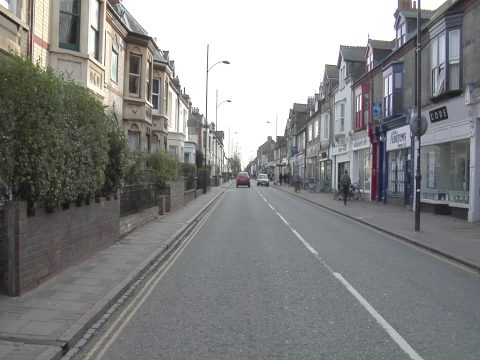}\label{fig:img2}}\hfill
\subfigure{\includegraphics[width=0.27\textwidth]{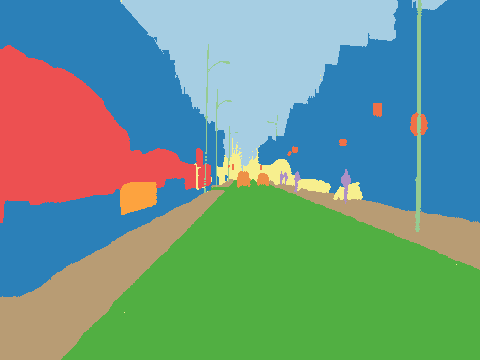}\label{fig:GT2}}\hfill 
\subfigure{\includegraphics[width=0.27\textwidth]{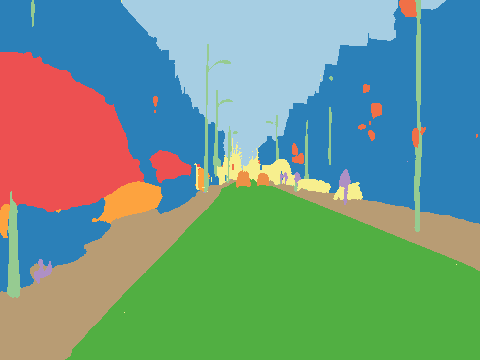}\label{fig:pred2}}\hfill \\
\subfigure{\includegraphics[width=0.27\textwidth]{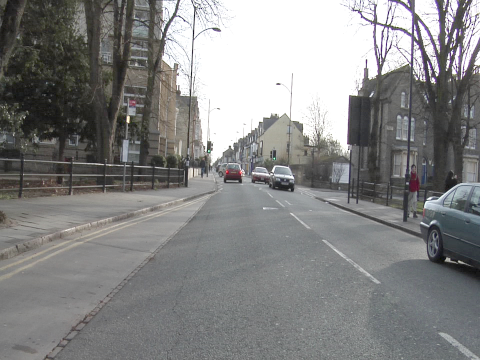}\label{fig:img3}}\hfill
\subfigure{\includegraphics[width=0.27\textwidth]{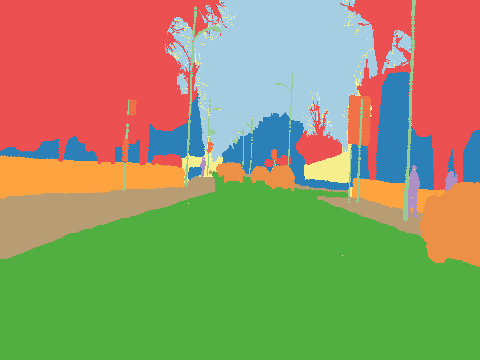}\label{fig:GT3}}\hfill
\subfigure{\includegraphics[width=0.27\textwidth]{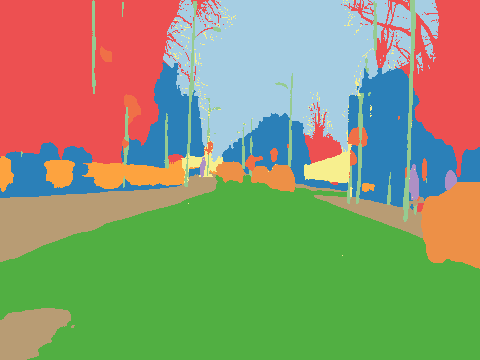}\label{fig:pred3}}\hfill \\
\subfigure{\includegraphics[width=0.27\textwidth]{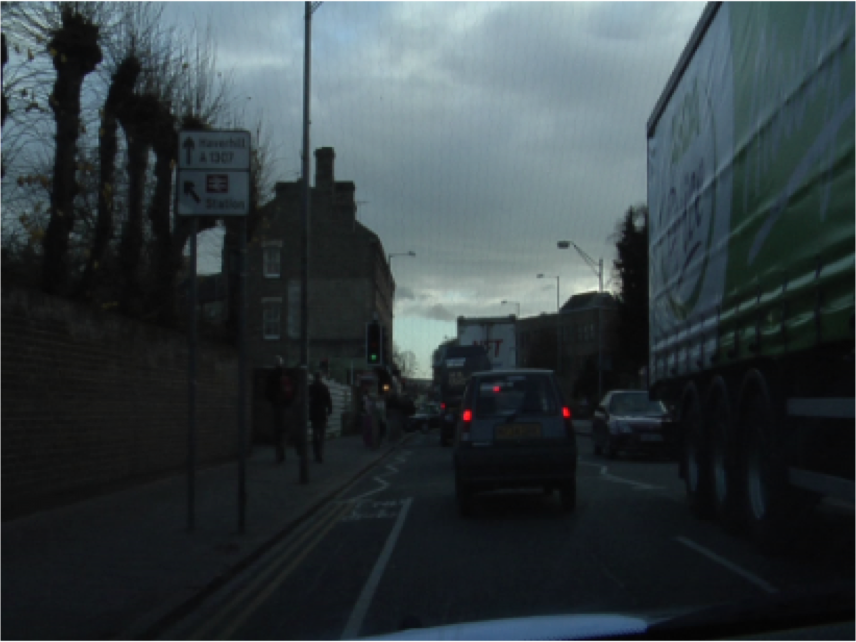}\label{fig:img4}}\hfill
\subfigure{\includegraphics[width=0.27\textwidth]{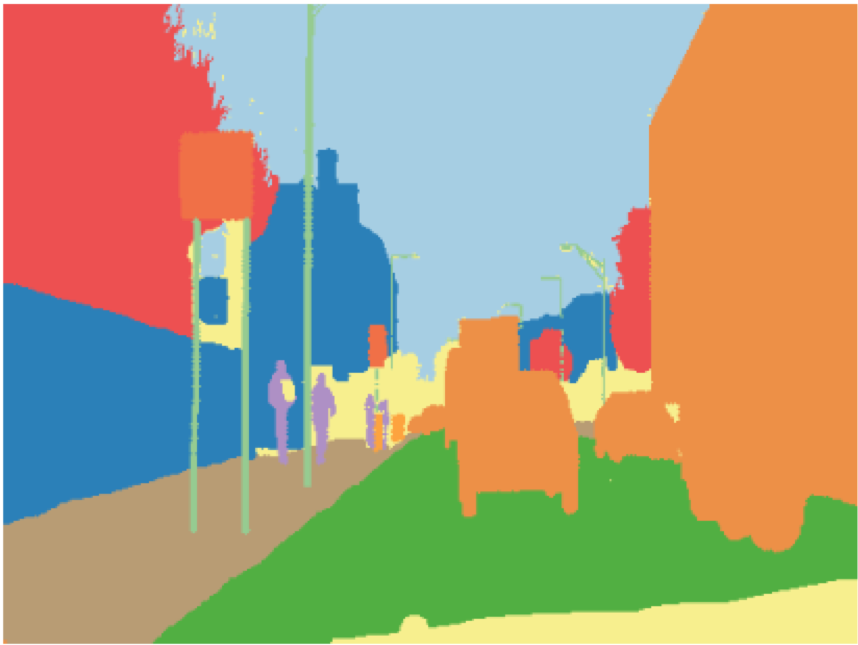}\label{fig:GT4}}\hfill
\subfigure{\includegraphics[width=0.27\textwidth]{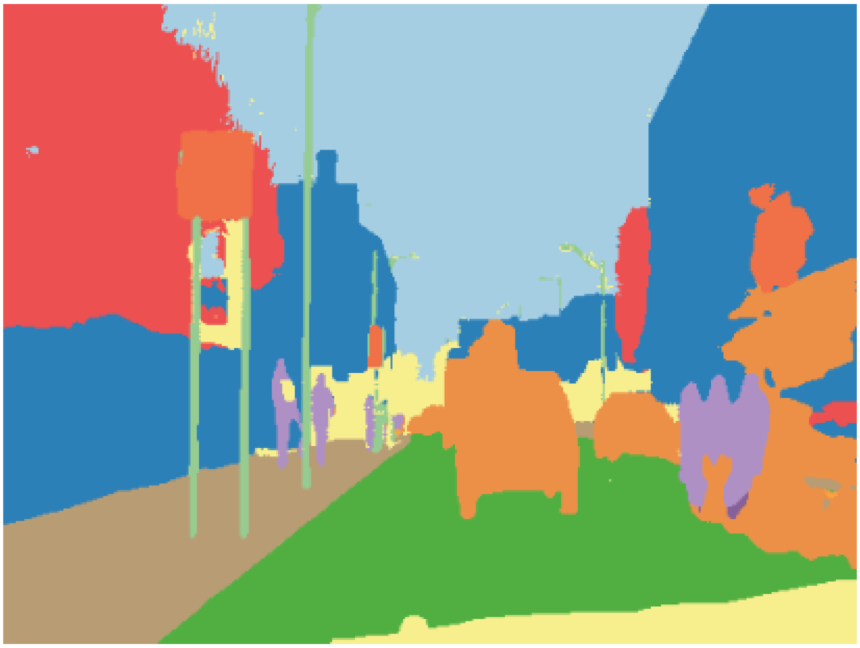}\label{fig:pred4}}\hfill\\
\subfigure{\includegraphics[width=0.27\textwidth]{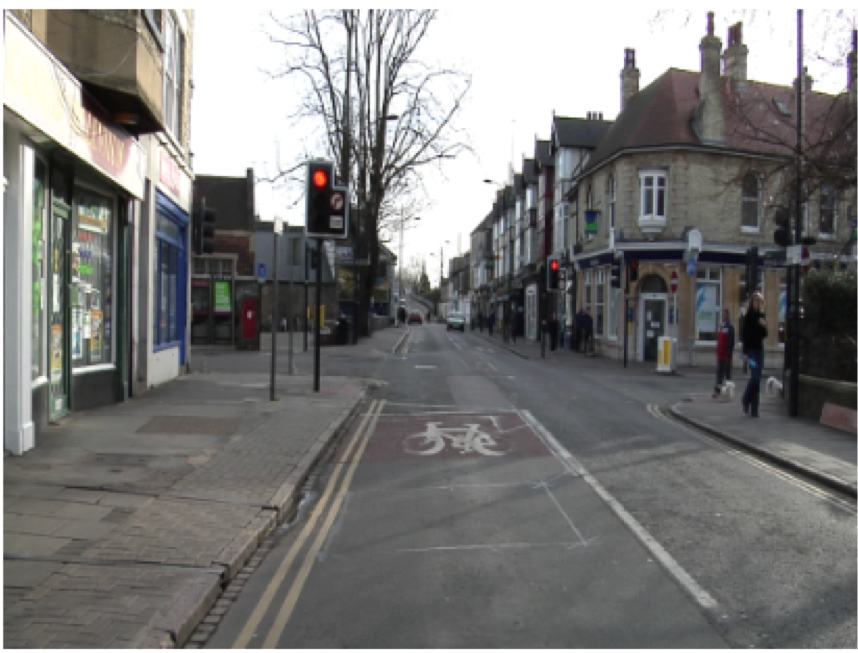}\label{fig:img5}}\hfill
\subfigure{\includegraphics[width=0.27\textwidth]{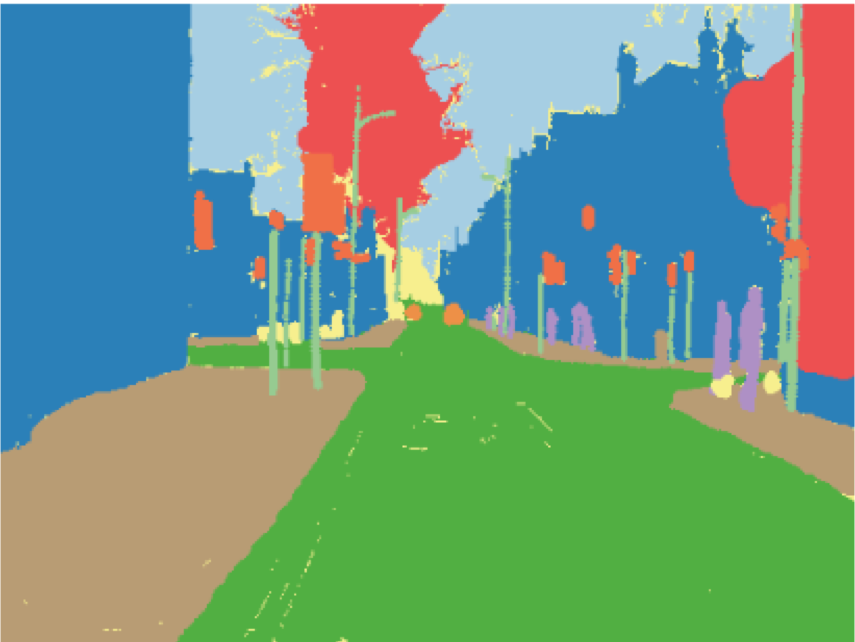}\label{fig:GT5}}\hfill
\subfigure{\includegraphics[width=0.27\textwidth]{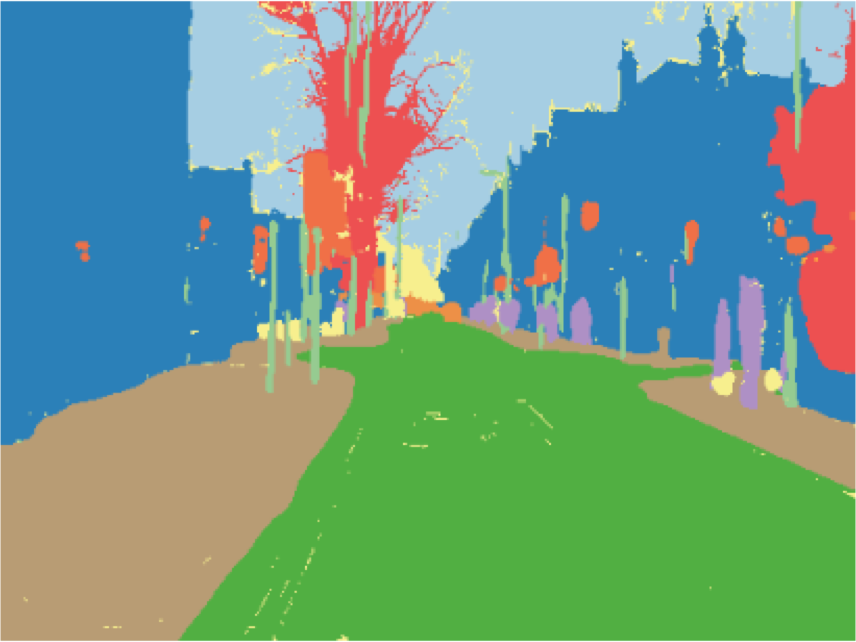}\label{fig:pred5}}\hfill\\
\caption{Qualitative results on the CamVid test set. Pixels labeled in yellow are void class. Each row represents (from left to right): original image, original annotation (ground truth) and prediction of our model.}
\label{fig:predictions}
\end{figure*}

Figure \ref{fig:predictions} shows some qualitative segmentation results on the CamVid dataset. Qualitative results are well aligned with the quantitative ones, showing sharp segmentations that account for a lot of details. For example, trees, column poles, sidewalk and pedestrians appear very well sketched. Among common errors, we find that thin details found in trees can be confused with column poles (see fifth row), buses and trucks can be confused with buildings (fourth row), and shop signs can be confused with road signs (second row).  

\begin{table*}
\footnotesize
\centering
\setlength\tabcolsep{5.3pt} 
\begin{minipage}{\textwidth} 
 \begin{tabular}{c | c | c || c | c | c | c | c | c | c | c | c | c | c || c | c } 
 
Model & \rotatebox{90}{Pretrained}& \rotatebox{90}{\# parameters (M)} &\rotatebox{90}{Building} & \rotatebox{90}{Tree} & \rotatebox{90}{Sky} & \rotatebox{90}{Car} & \rotatebox{90}{Sign} & \rotatebox{90}{Road} & \rotatebox{90}{Pedestrian} & \rotatebox{90}{Fence} & \rotatebox{90}{Pole} & \rotatebox{90}{Sidewalk} & \rotatebox{90}{Cyclist} & \rotatebox{90}{Mean IoU} & \rotatebox{90}{Global accuracy}  \\ 
\hline \hline
SegNet  \cite{SegNet2015}   & \Checkmark & $29.5$ & $68.7$ & $52.0$ & $87.0$ & $58.5$ & $13.4$ & $86.2$ & $25.3$ & $17.9$ & $16.0$ & $60.5$ & $24.8$ & $46.4$ & $62.5$\\ 
\hline
Bayesian SegNet \cite{KendallBC15}      & \Checkmark & $29.5$ & \multicolumn{11}{|c||}{n/a} &  $63.1$ & $86.9$\\
\hline
DeconvNet \cite{noh2015learning} & \Checkmark & $252$ & \multicolumn{11}{|c||}{n/a} &  $48.9$ & $85.9$\\
\hline
Visin et al. \cite{VisinKCBMC15} & \Checkmark & 32.3 & \multicolumn{11}{|c||}{n/a} & $58.8$ & $88.7$\\
\hline
FCN8 \cite{long2015fully} & \Checkmark & $134.5$ & $77.8$ & $71.0$ & $88.7$ & $76.1$ & $32.7$ & $91.2$ & $41.7$ & $24.4$ & $19.9$ & $72.7$ & $31.0$ & $57.0$ & $88.0$\\ 
\hline
DeepLab-LFOV \cite{chen14semantic} & \Checkmark & 37.3 & $81.5$ & $74.6$ & $89.0$ & $82.2$ & $42.3$ & $92.2$ & $48.4$ & $27.2$ & $14.3$ & $75.4$ & $50.1$ & $61.6$ & $-$\\ 
\hline
Dilation8  \cite{YuKoltun2016}  & \Checkmark  & $140.8$ & $82.6$ & $76.2$ & $89.0$ & $84.0$ & $46.9$ & $92.2$ & $56.3$ & $35.8$ & $23.4$ & $75.3$ & $55.5$ & $65.3$ & $79.0$\\ 
\hline 
Dilation8 + FSO \cite{KunduCVPR16} & \Checkmark  & $140.8$ & $\textbf{84.0}$ & $77.2$ & $91.3$ & $\textbf{85.6}$ & $\textbf{49.9}$ & $92.5$ & $59.1$ & $\textbf{37.6}$ & $16.9$ & $76.0$ & $\textbf{57.2}$ & $66.1$ & $88.3$\\ 
\hline  \hline
Classic Upsampling  & \ding{55} & $20$ & $73.5$ & $72.2$ & $92.4$ & $66.2$ & $26.9$ & $90.0$ & $37.7$ & $22.7$ & $30.8$ & $69.6$ & $25.1$ & $55.2$ & $86.8$ \\
\hline
FC-DenseNet56 (k=12) & \ding{55} & $1.5$ & $77.6$ & $72.0$ & $92.4$ & $73.2$ & $31.8$ & $92.8$ & $37.9$ & $26.2$ & $32.6$ & $79.9$ & $31.1$ & $58.9$ & $88.9$\\
\hline
FC-DenseNet67 (k=16) & \ding{55} & $3.5$ & $80.2$ & $75.4$ & $93.0$ & $78.2$ & $40.9$ & $\textbf{94.7}$ & $58.4$ & $30.7$ & $\textbf{38.4}$ & $81.9$ & $52.1$ & $65.8$ & $90.8$ \\
\hline
FC-DenseNet103 (k=16)& \ding{55} & $9.4$ & $83.0$ & $\textbf{77.3}$ & $\textbf{93.0}$ & $77.3$ & $43.9$ & $94.5$ & $\textbf{59.6}$ & $37.1$ & $37.8$ & $\textbf{82.2}$ & $50.5$ & $\textbf{66.9}$ & $\textbf{91.5}$\\
\hline

 \end{tabular}
 \end{minipage}
 \vspace{0.2cm}
 \caption{Results on CamVid dataset. Note that we trained our own pretrained FCN8 model}
 \label{tab:CamVid}
 \setlength\tabcolsep{6pt} 
\end{table*}

\subsection{Gatech dataset} 
Gatech\footnote{http://www.cc.gatech.edu/cpl/projects/videogeometriccontext/} \cite{VideoGeometricContext2013} is a geometric scene understanding dataset, which consists of 63 videos for training/validation and 38 for testing. Each video has between 50 and 300 frames (with an average of 190). A pixel-wise segmentation map is provided for each frame. There are 8 classes in the dataset: \textit{sky}, \textit{ground}, \textit{buildings}, \textit{porous} (mainly trees), \textit{humans}, \textit{cars}, \textit{vertical mix} and \textit{main mix}. The dataset was originally built to learn 3D geometric structure of outdoor video scenes and the standard metric for this dataset is mean global accuracy.

We used the FC-DenseNet103 model pretrained on CamVid, removed the softmax layer, and finetuned it for 10 epochs with crops of $224 \times 224$ and batch size 5. Given the high redundancy in Gatech frames, we used \emph{only} one out of 10 frames to train the model and tested it on all full resolution test set frames. 

In Table \ref{tab:GATECH}, we report the obtained results. We compare the results to the recently proposed method for video segmentation of \cite{Tran16v2v}, which reports results of their architecture with 2D and 3D convolutions. Frame-based 2D convolutions do not have temporal information. As it can be seen in Table \ref{tab:GATECH}, our method gives an impressive improvement of 23.7\% in global accuracy with respect to previously published state-of-the-art with 2D convolutions. Moreover, our model (trained with only 2D convolutions) also achieves a significant improvement over state-of-the-art models based on spatio-temporal 3D convolutions ($3.4\%$ improvement).

\begin{table}
\centering
 \begin{tabular}{c || c }
 Model & Acc. \\
 \hline \hline
 \multicolumn{2}{l}{\emph{2D models (no time)}} \\
 \hline 
 2D-V2V-from scratch \cite{Tran16v2v} & $55.7$\\ \hline
  FC-DenseNet103 & $\mathbf{79.4}$\\ 
 \hline 
 \hline
 \multicolumn{2}{l}{\emph{3D models (incorporate time)}} \\
 \hline
 3D-V2V-from scratch \cite{Tran16v2v} & $66.7$\\
 \hline
 3D-V2V-pretrained \cite{Tran16v2v} & $76.0$\\
 \hline

 \end{tabular}
 \vspace{0.2cm}
 \caption{Results on Gatech dataset}
 \label{tab:GATECH}
\end{table}

\section{Discussion}

Our fully convolutional DenseNet implicitly inherits the advantages of DenseNets, namely: (1) parameter efficiency, as our network has substantially less parameters than other segmentation architectures published for the Camvid dataset; (2) implicit deep supervision, we tried including additional levels of supervision to different layers of our network without noticeable change in performance; and (3) feature reuse, as all layers can easily access their preceding layers not only due to the iterative concatenation of feature maps in a dense block but also thanks to skip connections that enforce connectivity between downsampling and upsampling path.

Recent evidence suggest that ResNets behave like ensemble of relatively shallow networks \cite{VeitWB16}: "Residual networks avoid the vanishing gradient problem by introducing short paths which can carry gradient throughout the extent of very deep networks". It would be interesting to revisit this finding in the context of fully convolutional DenseNets. Due to iterative feature map concatenation in the dense block, the gradients are forced to be passed through networks of different depth (with different numbers of non-linearities). Thus, thanks to the smart connectivity patterns, FC-DenseNets might represent an ensemble of variable depth networks. This particular ensemble behavior would be very interesting for semantic segmentation models, where the ensemble of different paths throughout the model would capture the multi-scale appearance of objects in urban scene.

\section{Conclusion}
In this paper, we have extended DenseNets and made them fully convolutional to tackle the problem semantic image segmentation. The main idea behind DenseNets is captured in dense blocks that perform iterative concatenation of feature maps. We designed an upsampling path mitigating the linear growth of feature maps that would appear in a naive extension of DenseNets.

The resulting network is \emph{very deep} (from $56$ to $103$ layers) and has \emph{very few} parameters, about 10 fold reduction w.r.t. state-of-the-art models. Moreover, it improves state-of-the-art performance on challenging urban scene understanding datasets (CamVid and Gatech), without neither additional post-processing, pretraining, nor including temporal information.

\textbf{Aknowledgements} \\

The authors would like to thank the developers of Theano \cite{Theano} and Lasagne \cite{lasagne}. Special thanks to Fr\'{e}d\'{e}ric Bastien for his work assessing the compilation issues. Thanks to Francesco Visin for his well designed data-loader \cite{Visin}, as well as Harm de Vries for his support in network parallelization, and Tristan Sylvain. We acknowledge the support of the following agencies for research funding and computing support: Imagia Inc., Spanish projects TRA2014-57088-C2-1-R \& 2014-SGR-1506, TECNIOspring-FP7-ACCIÓ grant.

{\small
\bibliographystyle{ieee}
\bibliography{egbib}
}

\end{document}